%% file: root.tex
\documentclass[letterpaper, 10 pt, conference]{ieeeconf}  

\IEEEoverridecommandlockouts                              

\usepackage{amsmath}
\usepackage{xspace}

\usepackage[utf8]{inputenc} 
\usepackage[T1]{fontenc}    
\usepackage{hyperref}       
\usepackage{url}            
\usepackage{booktabs}       
\usepackage{amsfonts}       
\usepackage{nicefrac}       
\usepackage{microtype}      
\usepackage{graphicx}
\usepackage{amssymb}
\usepackage{wrapfig}
\usepackage{dsfont}
\usepackage[ruled, noend]{algorithm2e}
\usepackage{mathrsfs}
\usepackage{multicol}
\usepackage{threeparttable}
\usepackage{multirow}
\usepackage{comment}
\let\labelindent\relax
\usepackage{enumitem}
\usepackage{bm}
\usepackage{pifont}
\usepackage{threeparttable}
\usepackage{dsfont}
\usepackage{makecell}
\usepackage[dvipsnames, table]{xcolor}
\usepackage{caption}
\usepackage{subcaption}
\usepackage[export]{adjustbox}
\usepackage{tcolorbox}
\usepackage{cite}
\definecolor{darkgreen}{rgb}{0.0, 0.5, 0.0}
\definecolor{navyblue}{rgb}{0.0, 0.0, 0.5} 
\newcommand{\qa}[1]{{\textcolor{RoyalBlue}{{#1}}}}
\title{\LARGE \bf
Query-Centric Diffusion Policy for Generalizable Robotic Assembly
}

\author{Ziyi Xu$^{1*}$, Haohong Lin$^{1*}$, Shiqi Liu$^{1*}$, and Ding Zhao$^{1}$ 
\thanks{*Equal contribution.}%
\thanks{$^{1}$Department of Mechanical Engineering, Carnegie Mellon University, Pittsburgh, PA 15213 USA, {\tt\small \{ziyix2, haohongl, shiqiliu\}@andrew.cmu.edu}
}
\thanks{Project Page: \href{https://calcualatexzy.github.io/qdp-homepage/}{https://calcualatexzy.github.io/qdp-homepage/}}
}

\begin{document}

\maketitle
\thispagestyle{empty}
\pagestyle{empty}

\begin{abstract}
    The robotic assembly task poses a key challenge in building generalist robots due to the intrinsic complexity of part interactions and the sensitivity to noise perturbations in contact-rich settings. The assembly agent is typically designed in a hierarchical manner: high-level multi-part reasoning and low-level precise control. However, implementing such a hierarchical policy is challenging in practice due to the mismatch between high-level skill queries and low-level execution. To address this, we propose the Query-centric Diffusion Policy (QDP), a hierarchical framework that bridges high-level planning and low-level control by utilizing queries comprising objects, contact points, and skill information. QDP introduces a query-centric mechanism that identifies task-relevant components and uses them to guide low-level policies, leveraging point cloud observations to improve the policy's robustness. We conduct comprehensive experiments on the FurnitureBench in both simulation and real-world settings, demonstrating improved performance in skill precision and long-horizon success rate. In the challenging insertion and screwing tasks, QDP improves the skill-wise success rate by over 50\% compared to baselines without structured queries.

\end{abstract}

\begin{figure*}[htbp]
    \centering
    \includegraphics[width=0.98\linewidth]{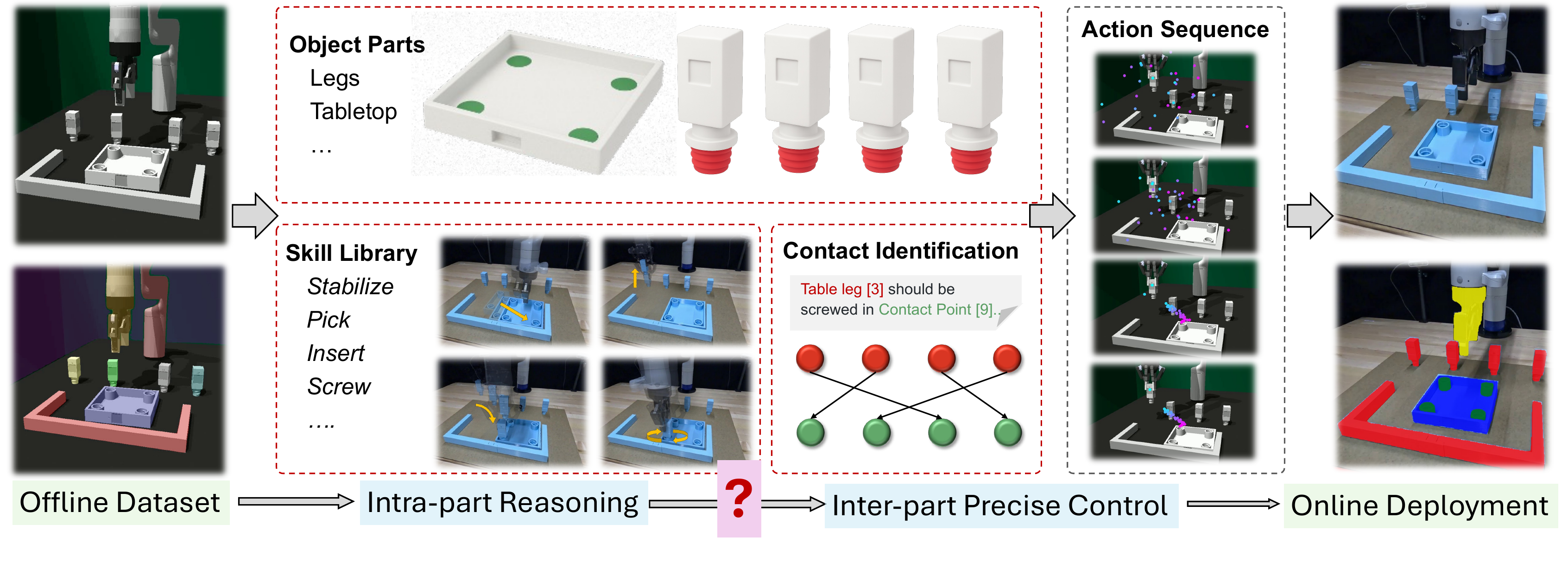}
    \vspace{-5mm}
    \caption{Overview of the furniture assembly task. The task nature comprises three parts: objects, contact points, and skill library. An effective assembly policy should identify the correct contact relationship given the assembly context and generate precise action sequences for the robot arm, bridging the gap between intra-part reasoning and inter-part precise control. }
    \label{fig:motivation}
\end{figure*}

\input{1_introduction}
\input{2_background}
\input{3_formulation}
\input{4_methodology}
\input{5_experiments}
\input{6_conclusions}



\bibliographystyle{IEEEtran}
\input{root.bbl}


\end{document}

%% file: 1_introduction.tex
\section{INTRODUCTION}

\label{sec:intro}
Contact-rich manipulation has been widely recognized as a critical task when building generalist intelligent robots~\cite{kim2024openvla, team2024octo, black2024pi0}. In this field, robot assembly~\cite{lee2021ikea, heo2023furniturebench} stands out because it requires policies that are both precise and versatile to control the robot arm and interact with multiple objects.
Despite the access to some offline demonstration data from human priors, robotic assembly poses two key challenges: inter-part relational reasoning and intra-part precise control in the online deployment. As is visualized in Figure~\ref{fig:motivation}, the first challenge arises from the long-horizon, multi-part nature of the task, which demands accurate prediction of the next skill based on current observations, as well as identifying the correct objects for interaction. The second challenge becomes especially difficult in contact-rich scenarios where successful assembly hinges on precise object alignment. Even minor noise or occlusions in raw sensory observations can completely fail the sim-to-real transfer of low-level \textit{non-prehensile} control policies.

Various methods have been explored to address these challenges. One line of work focuses on high-level reasoning by leveraging cross-modality affordance-based approaches~\cite{huang2023voxposer}, graph-based key point reasoning~\cite{huang2024rekep}, or skill-based retrieval~\cite{zhang2024extract, lin2024generalize}. However, these methods rely heavily on heuristic-based low-level controllers and predefined skill libraries, thus limiting their adaptability and precision.
Meanwhile, recent advances in robot learning have significantly improved the precision and adaptability of low-level policies, thanks to (i) powerful imitation learning backbones such as diffusion models~\cite{chi2023diffusion, janner2022planning, ajay2022conditional} and transformer-based architectures~\cite{chen2021decision, wang2024scaling}, (ii) enhanced learning regimes like residual policies~\cite{chi2024iterative} and safe failure prevention methods~\cite{ahmad2016safe}, and (iii) the integration of sensor modalities beyond vision and proprioception, such as tactile sensing~\cite{yu2024mimictouch, lin2024generalize} or point clouds~\cite{jiang2024transic} for improved contact modeling. 
Still, transferring these policies from simulation to the real world remains challenging due to the inherent difficulty of accurately simulating contact dynamics.

In addition to the individual challenges of high-level and low-level policy design, integrating these two levels in a hierarchical framework introduces further complications. For instance, high-level policies may mis-specify objects or skills, causing the low-level policy to rely on the \textit{false queries} that include irrelevant factors such as background pixels or non-impactful objects, ultimately leading to failure in real-world environments. Hence, it is crucial to establish a parsimonious proxy between high-level and low-level modules.

To address these challenges and establish a robust interface between high-level and low-level modules, we propose Query-centric Diffusion Policy (QDP). Our framework leverages powerful pre-trained foundation models to extract high-level information by specifying both the requisite skills and target objects as a query. This query then serves as a powerful precondition to guide point cloud-based low-level control, forcing the robot agent to focus on the current task-relevant components when generating the action chunks under different contexts.
Our contributions are threefold:

\begin{itemize}[leftmargin=0.2in]
    \item We introduce QDP, a query-centric diffusion policy framework that selects task-relevant queries for guiding low-level policies, enabling accurate skill selection and precise object interaction. 
    \item We propose a query-conditioned policy learning scheme to model the complex geometry captured by point cloud observations, improving precision and facilitating sim-to-real transfer. 
    \item We demonstrate the effectiveness of our approach on FurnitureBench in both simulation and real-world settings, showcasing robust performance under object misalignment and human intervention.
\end{itemize}

%% file: 2_background.tex
\section{RELATED WORK}
\label{sec:background}

\textbf{Sequential decision making for robotic assembly}
Behavior learning methods are promising for addressing the combinatorial challenges of assembly sequence planning and fine-grained manipulation in robotic assembly tasks. Prior works have tackled the former problem by designing feasible assembly graphs \cite{tian2024asap}, disassembling strategies \cite{tian2022assemble}, rearranging components through segmentation \cite{li2023rearrangement}, or even utilizing manual guidance \cite{wang2022ikea}. While the latter challenge of manipulation involves precise control and adaptability, recent studies centered around FurnitureBench have employed contact-rich manipulation~\cite{lin2024generalize}, residual policies bridging sim-to-real gap \cite{jiang2024transic}, \cite{ankile2024imitation}, fine-tuning diffusion policies \cite{ren2024diffusion}, or extracting skills from offline dataset \cite{zhang2024extract}. Our approach tackles the two challenges by developing a unified hierarchical decision-making framework for robotic assembly.

\textbf{Hierarchical imitation learning}
Instead of learning the entire task using a single policy, we can decompose the long-horizon assembly problem into reusable sub-tasks, thereby introducing a hierarchical learning framework. This typically involves a high-level policy and a set of low-level policies. On the high-level decision-making side, approaches often rely on the observable conditions \cite{driess2021learning} and accurate system dynamics \cite{chua2018deep} to realize the precondition of applying skill and its effects. On the low-level side, previous skill-based reinforcement learning methods benefit from generalization through skill discovery but face challenges in skill chaining \cite{lee2021adversarial} and reward design \cite{lin2024generalize}, especially in contact-rich environments. In this context, imitation learning can bridge the gaps between sub-tasks and avoid reward shaping by using full demonstration data. Previous data-driven hierarchical frameworks have included approaches such as sequence segmentation \cite{kipf2019compile}, human-in-the-loop data collection systems \cite{mandlekar2023human}, and high-level planning based on heuristics \cite{chitnis2016guided}. Our focus, however, is on designing a hinge connecting high-level and low-level, generalizing across different assets and orders.

\textbf{Generative diffusion model for planning}
Recent advancements in diffusion-based generative strategies have led to widespread adoption in robotics, particularly in imitation learning~\cite{chi2023diffusion} and planning~\cite{janner2022planning, ajay2022conditional}. These approaches have proven effective in capturing multi-modal distributions for planning and policy learning. Building on this diffusion-based framework, previous work explored learning visuomotor policy by integrating various conditions, including 3D representation~\cite{ze20243d}, equivariant models~\cite{yang2024equibot}, and affordance-based guidance~\cite{wu2024afforddp}, or performing hierarchical abstraction~\cite{li2023hierarchical, chen2024simple, ma2024hierarchical, liang2024skilldiffuser}. 

%% file: 3_formulation.tex
\section{PROBLEM FORMULATION}
\label{sec:formulation}
\paragraph{Task objective} 
Given a multi-part, long-horizon robot assembly task, our goal is to both generalize in policy-chaining and enhance the quality of robotic assembly, especially in bridging simulation and real-world environments. In general, the objective of our furniture assembly task contains improvements in two parts: higher level decision-making and lower level action-execution. Specifically, we aim to generate feasible high-level execution sequences and precise low-level policies based on 3D representations. 

\paragraph{Skills: motion primitives}
Following the setup in FurnitureBench \cite{heo2023furniturebench}, we divide the full assembly sequence to a set of skills \(\{\textit{stabilize, grasp, insert, screw}\}\), where \textit{stabilize} refers to pushing the tabletop to the corner of the stabilizer, \textit{grasp} is to pick up the objects with propose grasp pose, \textit{insert} requires inserting objects to the target contact point with extra precision, and \textit{screw} focuses on threading the table legs upright into the holes.

\paragraph{Objects: furniture parts} 
To equip robots with the capability to assemble a wide range of furniture, we classify furniture components into two major categories.
\begin{itemize}[leftmargin=0.2in]
\item \textbf{Primary Component}: These are the core structures of the furniture, such as tabletops, lamp bases, and chair seats.
\item \textbf{Secondary Components}: These parts complete or assist the furniture's function, including table legs, lamp bulbs, and drawer handles.
\end{itemize}
Additionally, we define \textbf{Contact Points} as specific locations on a Primary Component where a Secondary Component is intended to be attached.

\paragraph{Framework formulation}
Given the complexity of the task, we emphasize the importance of effective \textit{querying}. Assume we have a set of \( N_s \) skill primitives as sub-tasks, \( N_p \) furniture parts and \( N_c \) contact points available on the workbench, we define a query set \( \mathcal Q \) as:
\begin{equation}
   \mathcal{Q}
  \triangleq \Bigl\{\,q=(q_{ij}^{(\ell)})\in\{0,1\}^{N_s\times N_p\times N_c}
    \;\Bigm|\;
    \sum_{\ell=1}^{N_s}\sum_{i=1}^{N_p}\sum_{j=1}^{N_c}
      q_{ij}^{(\ell)}
    =1
  \Bigr\}. 
\end{equation}  
At every sub-task performance, our system generates a query \(q_{ij}^{(\ell)} \in \mathcal Q \), denoting a one-hot selection of furniture part \(i\), contact point \(j\) under the sub-task \( \ell \). 

In our task, the action space \( \mathcal{A} \) consists of the end-effector's delta poses and absolute poses. The state space \( \mathcal{S} \) includes the initial and goal assembly states of the full task, along with observation space $\mathcal{O}$ comprising RGB and depth inputs from external cameras. Additionally, it contains workspace-related states, including the states of individual furniture parts \( s_i \), contact points \( s_j \), and the end-effector \( s_e \).

Building on the notation above, we formulate our robot manipulation task as a complex, long-horizon, partially observable \textbf{query-centric} Task and Motion Planning (TAMP) problem~\cite{garrett2021integrated}. This formulation involves a query-centric execution sequence orderly obtained from \( \mathcal{Q} \), a series of motion planning procedures for each sub-task, and the initial and goal states drawn from \( \mathcal{S} \). Each motion planning procedure maps an observation input from \( \mathcal{O} \) to an action output in \( \mathcal{A} \).

\paragraph{Additional assumptions}
We assume smooth transitions between consecutive sub-tasks within the skill chaining process. We make our best efforts to ensure the terminal state of a given sub-task is a subset of the initial state of the subsequent sub-task.

%% file: 4_methodology.tex
\vspace{3mm}
\section{METHODOLOGY}
\begin{figure*}
    \centering
    \includegraphics[width=1.00\linewidth]{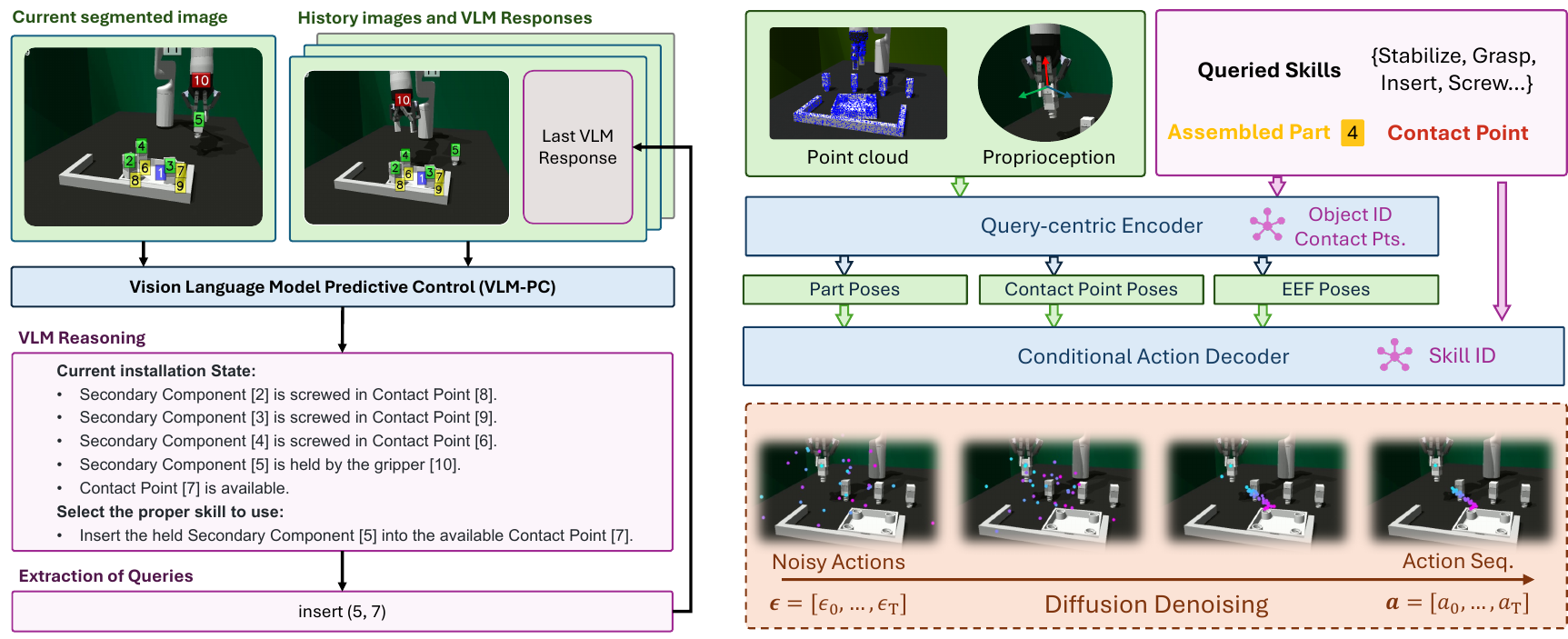}
    \caption{The proposed QDP framework adopts a hierarchical structure consisting of a high-level and a low-level policy. The high-level policy processes the initial RGB scene image, annotates it as SoM with SAM 2~\cite{ravi2024sam}, and leverages a Vision-Language Model (VLM) to identify key components. Using \textcolor{navyblue}{VLM Predictive Control (VLM-PC)}, it takes in the \textcolor{darkgreen}{current image} and \textcolor{darkgreen}{interaction history} to dynamically select skills and objects as \textcolor{purple}{queries}. 
    At the core of the system, the low-level policy takes in the \textcolor{purple}{queries} and operates directly on \textcolor{darkgreen}{point cloud observations} and \textcolor{darkgreen}{robot proprioception}. The \textcolor{navyblue}{query-centric encoder} explicitly estimates the poses of the queried components in the current context and generates precise action sequences of the end effector. By conditioning a \textcolor{navyblue}{diffusion policy} on these skill queries and assembly targets, the QDP generates fine-grained, context-aware action sequences. 
    }
    \label{fig:overview}
\end{figure*}
\label{sec:method}

In this section, we elaborate on the core design of our proposed query-centric diffusion policy. 
The methodology begins with the generation of queries from VLM-PC in the high-level part (Section~\ref{subsec:high_level}), followed by the introduction of query-centric diffusion policy training in the low-level phase (Section~\ref{subsec:low_level}). The entire pipeline of our method is illustrated in Figure~\ref{fig:overview}.

\subsection{High-level: Sequence Generator}
\label{subsec:high_level}
Our high-level policy aims to mitigate a long-horizon, contact-rich furniture assembly task by commanding the robot to sequentially apply low-level skills. 
By leveraging VLM Predictive Control (VLM-PC) \cite{chen2024commonsense}, our method dynamically selects appropriate low-level skills from the skill library based on the current scene image and the history of interactions.
The model incorporates a re-planning mechanism after each skill execution, allowing the system to handle external perturbations and changes in the environment effectively.

\paragraph{Components Identification} The high-level policy identifies furniture components from the initial scene image before installation. 
To improve the visual grounding capabilities of the VLM, which inherently lacks precise visual grounding \cite{zheng2024gpt}, we first segment the input image using SAM 2 \cite{ravi2024sam} and annotate it with markers generated by Set-of-Mark (SoM) \cite{yang2310set} at the beginning of the installation.
The annotated image is then processed by a VLM, which is prompted to identify markers for the primary components, secondary components, contact points, and the robotic arm gripper. 
Since the VLM relies on both the current image and the interaction history to maintain marker consistency across different steps, we track the initial segmentation masks throughout the sequence. 

\paragraph{Dynamic Skill Selection} The high-level policy dynamically determines the most suitable low-level skill to execute next. 
After completing each selected skill, the VLM is prompted to re-plan the installation based on the annotated current image and the previous $n$ interaction histories, including past annotated images and VLM outputs. 
The VLM first reasons the current installation progress, and identifies the state of each furniture component in case of external perturbations during installation. It is then prompted to select the most appropriate skill from the skill library. 

\subsection{Low-level: Query-conditioned Action Generator} 
\label{subsec:low_level}
We develop a query-centric low-level policy that generates an action chunk under the observation input and the selected query \(q_{ij}^{(\ell)}\) from the high-level policy. To bridge the sim-to-real gap, we propose to use a point cloud as the visual observation input. Numerical queries from the high-level sequence generator are embedded using a lookup table.
By using offline point cloud generated from mesh, we first train a query-centric encoder and explicitly supervise the output with pose of the queried furniture part and contact point. We then train an action decoder for each sub-task with the true state of the queried furniture part as input using the collected demonstrations in simulation under a diffusion policy framework.
Combined with the perception encoder, this forms a two-stage imitation learning structure for the low-level policy.

\paragraph{Query-centric Encoder}
The query-centric encoder is trained using the full-task dataset. To extract useful latent representations from the point clouds, we use a PointNet \cite{qi2017pointnet} architecture. The skill query \( \ell \) , furniture-part query \( i \) and contact point query \(j\) are numerically input into the encoder network through embeddings, concatenating the point cloud embedding. We also incorporate the end-effector states into a fully connected layer, hypothesizing that these end-effector states will help the encoder network focus on the specific furniture parts being manipulated at the current given moment. These features are concatenated and fused through an MLP network with dropouts, and the resulting representation is used to predict the estimated states. 3D point clouds without norms struggle with accurate rotation estimation. Hence, we assume the furniture parts are rotation invariant in z-axis, where the tabletop is constrained to rotate within the table plane, retaining only two euler angles representing rotation. The Mean Square Error (MSE) training loss \(L_{\text{EST}}\) is defined as a weighted mean squared error over the object-centric position \( P_i \) , rotation  \( R_i \) and the contact-centric position \( P_j \), where \(\hat{P}_i\), \(\hat{R}_i\) and \(\hat{P}_j\) denote the predicted values:
\begin{equation}
L_{\text{EST}} \! = \! \sum_{q_{ij}^{(\ell)} \in \mathcal Q} q_{ij}^{(\ell)} \cdot \big( \alpha   \| \hat{P}_i - P_i \|^2 \! + \beta  \| \hat{R}_i - R_i \|^2 \! + \gamma \| \hat{P}_j - P_j \|^2 \big),
\end{equation}
where $\alpha, \beta, \gamma$ are the coefficients of the weighted loss. 
The estimated poses, together with end-effector states, are also useful in sub-task transition, where we set terminal checks to chain the sub-tasks. Since this might fail in assembly check, a time-out check is set to prevent excessive steps of screwing.

\paragraph{Action Decoder} 
The full-task demonstrations are first divided according to the sub-task categories, and sub-task-specific policies are trained accordingly. The sub-task action decoder is built upon the Diffusion Policy framework \cite{chi2023diffusion}, leveraging a temporal CNN-based U-Net architecture with FiLM conditioning layers to effectively model and generate precise action chunks. With the intuition that action modality is more pronounced in position inputs, the denoising process is conditioned on the true state of the queried furniture part, the corresponding contact point, and the additional proprioception states of the end-effector. This allows the policy to learn to handle each sub-task \(\ell\) with the necessary precision and contextual awareness. The decoder output is in the end-effector operational space, further continued with an operational space controller (OSC) \cite{zhu2020robosuite}. Starting from a Gaussian noise \( \mathbf{a}^K \), the denoising network \( \mathbf{\epsilon}_{\theta}^{(\ell)} \) with parameters \( \theta \) performs \( K \) iterations to gradually denoise a random noise \( \mathbf{a}^K \) into the noise-free action chunk \( \mathbf{a}^0 \), processed in the following equation:
\begin{equation}
    \mathbf{a}^{k-1} = \alpha_k \left ( \mathbf{a}^{k} - \gamma_k \sum_{i=1}^{N_p} \sum_{j=1}^{N_c} q_{ij}^{(\ell)} \cdot \mathbf{\epsilon}_\theta (\mathbf{a}^{k}, k, \mathbf s) \right ) + \sigma_k \epsilon,
\end{equation}
where $\mathbf{s} = [s_i, s_j, s_e]$ is the queried state from high-level policy, \( \epsilon\sim \mathcal{N}(0, \mathbf{I}) \) is the Gaussian noise, and parameterized function \( \alpha_k, \gamma_k \) and \( \sigma_k \) are generated by the noise scheduler. The combinatorial Mean Square Error (MSE) training loss \( L_{ACT} \) for the action generator policy is: 
\begin{equation}
    L_{\text{ACT}} = \| \epsilon^k - \sum_{q_{ij}^{(\ell)} \in \mathcal Q} q_{ij}^{(\ell)} \cdot \mathbf{\epsilon}_{\theta}^{(\ell)} (\mathbf{a}^{0}+\epsilon^k, k, \mathbf{s} )  \|^2
\end{equation}

\subsection{Demonstration Data Collection and Annotation}
We collected full-task demonstrations finishing the whole furniture assembly sequence using a heuristic policy with Finite State Machine (FSM) in simulation. During every step of the collection, we mark the queried furniture parts and contact points in a numerical order, annotate sub-task transitions, and record the full states of the assets. To improve generalization, we introduce a jitter at each step, applying it uniformly to the states of every asset involved. For the furniture assets, we use a synthetic point cloud that is subsampled from mesh files, eliminating the need for prolonged rendering in simulation environments. For the robot arm, we mitigate the sim-to-real gap by collecting real-world point clouds of the gripper in both \textit{Open} and \textit{Close} modes, which replace the synthetic point clouds. In the real-world deployment, policies are applied zero-shot, with no additional data collection required. Every demonstration is not only initialized with a \textit{Medium} randomness~\cite{heo2023furniturebench}, but also has a randomly sampled execution order of the furniture parts, representing different sequences generated by the high-level sequence generator. We collect a total of 300 full-task demonstrations of the square table with randomly generated execution sequences in simulation. Given the high step count involved in the screw sub-task, we limit each demonstration of screwing to 50 steps, combining these with other sub-tasks to train the query-centric encoder. 

%% file: 5_experiments.tex
\section{EXPERIMENTS}
\label{sec:experiments}

\begin{figure*}[htbp]
  \centering
  \includegraphics[width=1\linewidth]{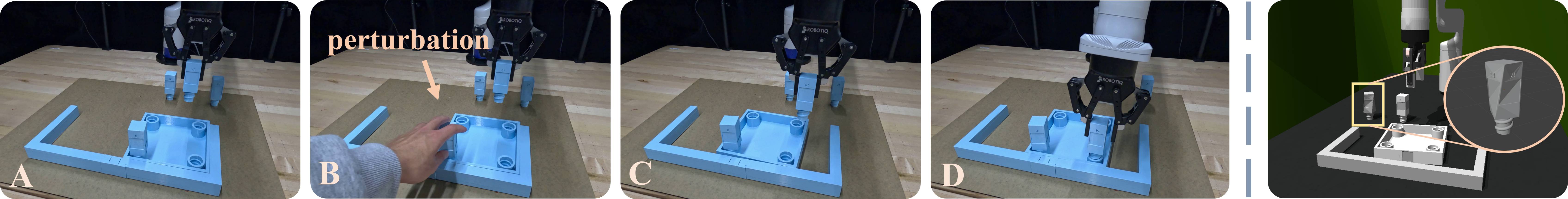}
  \caption{Robustness under perturbation: disturbance of table-top in real (left) and shape transformation in sim (right). }
  \label{fig:robustness}
\end{figure*}

\begin{figure*}[htbp]
  \centering
  \includegraphics[width=1\linewidth]{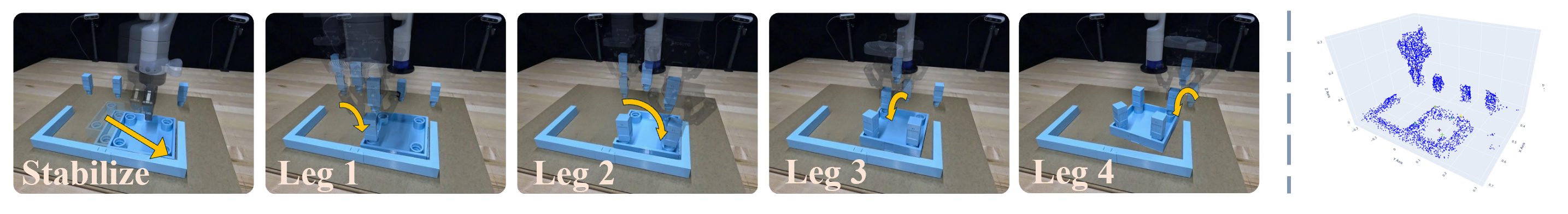}
  \caption{Step-by-step assembly process of QDP (left) and a visualization of the real-world point cloud (right).}
  \label{fig:real_skill_timelapes}
\end{figure*}

\subsection{Environment Design} 

\paragraph{Tasks Design}
We demonstrate our pipeline on the \textit{Square Table} assembly task, which involves attaching four table legs to the tabletop. This task consists of three sub-tasks for each leg assembly: \textit{grasp}, \textit{insert}, and \textit{screw}, along with an additional \textit{stabilize} step. The ultimate goal is to integrate all four legs onto the tabletop in a feasible assembly sequence, which requires long-horizon decision-making as well as stability and accuracy in both state estimation and action generation.

\begin{figure}[htbp]
  \centering
  \includegraphics[width=1.0\linewidth]{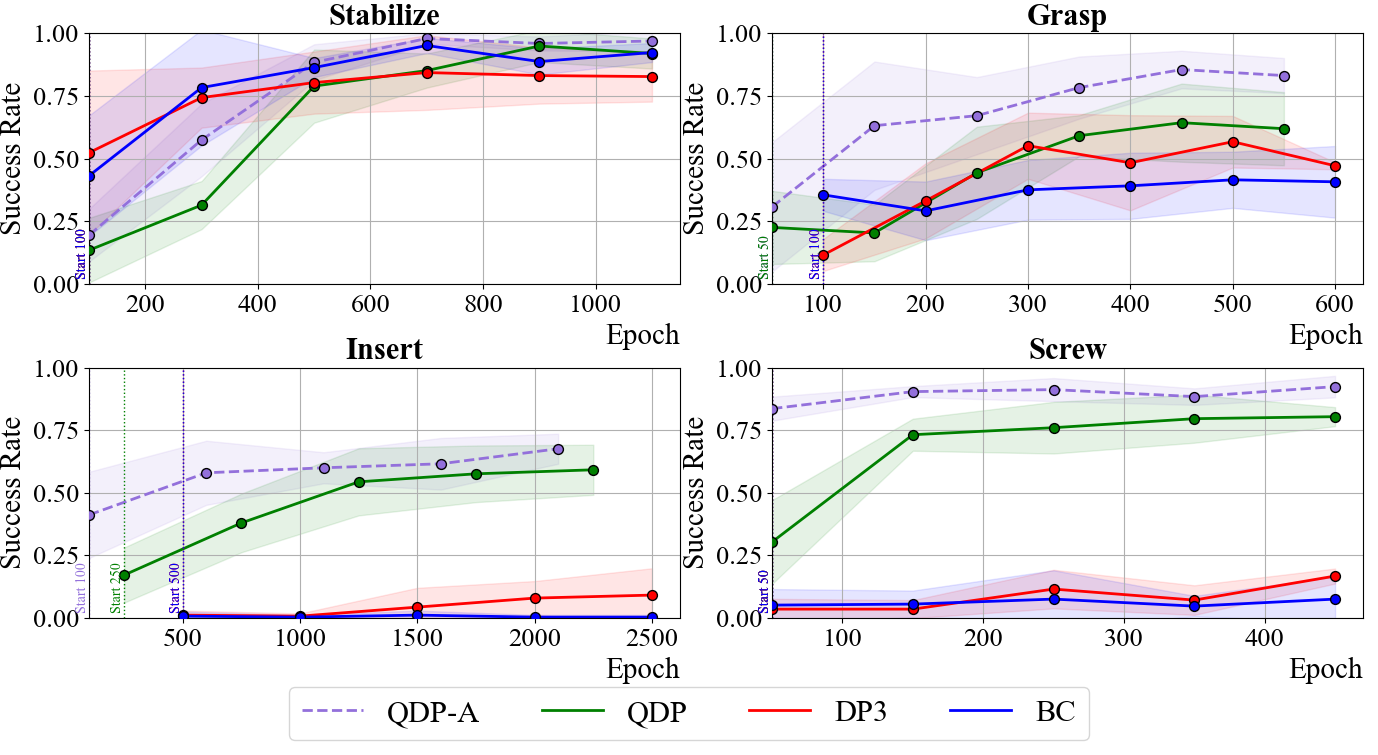}
  \caption{Learning curves for comparison of four methods: \textbf{QDP}, QDP with only the sub-task action decoder (\textbf{QDP-A}), \textbf{DP3} and \textbf{BC}. \textbf{QDP} demonstrates a better performance and faster convergence especially in bottleneck sub-tasks such as \textit{insert} and \textit{screw}.}
  \label{fig:evaluation}
\end{figure}

For numerical evaluation, we prefer using the \textit{One Leg} assembly task, containing all four basic sub-tasks as skill primitives. Notably, we define the success of the \textit{grasp} not only by successfully lifting the leg, but also by ensuring that the grasp occurs within a small, controlled area to prevent dangerous grips or potential collisions with the other legs.
We deploy our framework in simulation using Isaac Gym~\cite{makoviychuk2021isaac}, and validate it in the real world with a furniture assembly setup on the Kinova Gen3 robotic arm. 

\paragraph{Real-world Deployment}
\label{appendix:hardware}
We deploy our framework on the Kinova Gen3 collaborative robot arm, using four RealSense D435 cameras to capture real-world point clouds and video frames. The furniture parts are reset to a relatively fixed state, with randomness introduced in their position and rotation. We use a paperboard to fix the position of the \textit{stabilizer} relative to the robot arm. To avoid singularities during grasping, the furniture legs are repositioned to a standing position. However, we argue that this setup does not simplify the task, as a random yaw rotation is also applied, adding complexity to the scenario.

For real-world applications, we first integrate point clouds captured by all four cameras in the world frame and then clip out the workspace. Since the synthetic point clouds collected for training in simulation are generated based on mesh surface density, it is essential to preprocess the real-world point clouds through voxel down-sampling. Additionally, we segment the ground plane and filter out any outliers. After a final down-sampling step, we retain a total of 4096 points, as is illustrated in Figure~\ref{fig:real_skill_timelapes}(f).   

Our policy outputs an action chunk consisting of end-effector poses. To satisfy the quasi-static assumption, the Kinova Gen3 robot arm tracks the first eight points in the action chunk orderly, naturally resulting in longer motion durations, while model inference time is tested comparable to that of the original Diffusion Policy~\cite{chi2023diffusion}. 

\subsection{Low-level Evaluation Protocol}
We evaluate low-level QDP with two common IL baselines including Behavior Cloning (\textbf{BC}) and \textbf{DP3}~\cite{ze20243d}. For a fair comparison, we implement the same number of demonstrations, same PointNet backbone and evaluate per-skill performance among them. We focus on executing a fixed leg, as the baseline methods tend to underperform in a multi-component assembly setting. Since QDP has two learning stages, in order to fairly compare the convergence, we pair intermediate checkpoints of pose estimation with intermediate checkpoints of action decoder. The learning curves are shown in Figure~\ref{fig:evaluation}. 

For low-level QDP, we perform two ablation studies: 1) Removing the rotation invariance assumption (\textbf{w/o rot-inv}), where instead of using two euler angles representing rotation, we use full quaternions. 2) Removing the query-centric structure (\textbf{w/o query}), where instead of using the queried states, we estimate the states of all furniture parts jointly and choose the queried part heuristically. 

We also perform three breakdown studies: 1) We introduce a shape transformation by twisting the furniture leg (\textbf{v/ shape}: shape variation, as shown in Figure~\ref{fig:robustness}), which challenges the generalization of the state estimation module with point cloud input. 2) To ensure numerically valid results, we evaluate the \textit{One Leg} task on the final leg to be assembled, randomly varying the assembly order for the last hole (\textbf{v/ order}: order variation). For example, we randomly arrange the hole closest to the camera to be the last one assembled, while the other holes have already been completed. 3) Building on the \textit{One Leg} setup, we introduce an additional variation where we perturb the tabletop randomly when executing each sub-task (\textbf{v/ tabletop}: tabletop variation). This setup is conducted both in simulation and real-world, designed to validate the robustness of low-level skills. 

For the baselines, ablation study, and shape variation in the breakdown study, each evaluation is conducted 5 times, each 50 roll-outs in the simulation environment, using 10 parallel environments and randomly generating an executable order during the reset process. Other evaluations are conducted 5 times, each 10 roll-outs in single simulation environments, with the setup tailored to their specific configuration.

\subsection{High-level Evaluation Protocol}
The ablation study focused on evaluating the performance of QDP’s high-level variants: 1) QDP without VLM-PC (\textbf{w/o VLM-PC}), where the installation sequence is generated solely based on the initial frame, without utilizing VLM Predictive Control.
2) QDP without marker (\textbf{w/o Marker}), this variant excludes the markers generated by SoM in the input image.
3) QDP without interaction history (\textbf{w/o History}), this variant excludes interaction history input to the VLM.
Additionally, we analyzed the impact of history length on the policy’s performance. 

To evaluate the performance of the high-level policy, the evaluation procedure is set independently from the low-level policy since a correctly chosen skill might fail due to the inherent randomness of the low-level policy. In cases where the chosen skill failed during execution, we reset the simulated environment to its state before execution and attempted to re-execute the skill. If the skill could not be successfully executed within a predefined retry limit, it was considered a failure. Otherwise, we proceeded to evaluate the next chosen skill. 
We utilized GPT-4o as the VLM for the evaluation process.

\subsection{Results and Analysis}

In the following part, we answer the following research questions: 
\begin{itemize}[leftmargin=0.2in]
    \item \qa{\textbf{RQ1}}: How is QDP's precision in both simulation and the real furniture assembly tasks?
    \item \qa{\textbf{RQ2}}: How does each component of QDP contribute to its superior performance?
    \item \qa{\textbf{RQ3}}: How is the robustness of QDP against multifarious external perturbations?
\end{itemize}

\begin{table}[htbp]
    \centering
    \caption{Success rate of low-level ablations}
    \label{tab:simulation_results}
    \setlength{\tabcolsep}{4pt} 
    \begin{adjustbox}{max width=\linewidth}
    \begin{tabular}{@{}lcccc@{}}
        \toprule
        \textbf{Method} & \textit{stabilize} & \textit{grasp} & \textit{insert} & \textit{screw} \\
        \midrule
        \textbf{BC}           & $0.92 \pm 0.04$ & $0.41 \pm 0.14$ & $0.00 \pm 0.01$ & $0.08 \pm 0.09$ \\
        \textbf{DP3}           & $0.83 \pm 0.10$ & $0.47 \pm 0.02$ & $0.09 \pm 0.11$ & $0.17 \pm 0.03$ \\
        \textbf{QDP}           & $0.95 \pm 0.06$ & $0.64 \pm 0.14$ & $\bm{0.59 \pm 0.09}$ & $\bm{0.80 \pm 0.03}$ \\ \midrule
        \textbf{QDP w/o query}    & $\bm{0.96 \pm 0.03}$ & $0.60 \pm 0.10$ & $0.06 \pm 0.05$ & $0.33 \pm 0.09$ \\
        \textbf{QDP w/o rot-inv}      & $0.96 \pm 0.05$ & $\bm{0.70 \pm 0.08}$ & $0.50 \pm 0.08$ & $0.42 \pm 0.13$ \\ 
        \midrule
        \textbf{QDP v/ shape}     & $0.95 \pm 0.04$ & $0.56 \pm 0.11$ & $0.56 \pm 0.12$ & $0.76 \pm 0.10$ \\
        \textbf{QDP v/ order} & --- & $0.38 \pm 0.07$ & $0.28 \pm 0.13$ & $0.62 \pm 0.10 $ \\
        \textbf{QDP v/ tabletop} & $0.88 \pm 0.08 $ & $0.68 \pm 0.08$ & $0.46 \pm 0.11$ & ---\\
        \bottomrule
    \end{tabular}
    \end{adjustbox}
\end{table}

For~\qa{\textbf{RQ1}}, the overall success rate of QDP in performing the \textit{One Leg} task in simulation is 32\%. 
As shown in Figure~\ref{fig:evaluation} and Table~\ref{tab:simulation_results}, our query-centric structure outperforms the baselines especially in the \textit{insert} and \textit{screw} sub-task. Additionally, our framework shows a faster convergence in training low-level policies compared with baseline methods. It is evident that jointly estimating all furniture parts without the query structure results in a significant performance drop in the \textit{insert} task. This decline can be attributed to the improved attention provided by the query-centric approach, particularly in focusing on the current executing furniture part. In contrast, a better symmetry representation enhances performance in the \textit{screw} sub-task. This is because the screw task demands precise relocation, and the full quaternion representation may introduce ambiguity, hindering the network's interpretation. In the real-world deployment, QDP successfully completed the \textit{One Leg} task in a zero-shot manner, achieving a success rate of 7/18. The entire assembly process completed by QDP is illustrated in Figure~\ref{fig:real_skill_timelapes}(a)-(e).

For~\qa{\textbf{RQ2}}, we evaluate the importance of each design module in QDP low-level policy with two variants: \textbf{QDP w/o query} and \textbf{QDP w/o rot-inv}. We illustrate in Table~\ref{tab:simulation_results} the success rate of each skill in the assembly process with these ablation variants in the simulation environments. The results show that the structured queries and rotation invariance assumptions have a greater impact on the low-level policy in the harder task, such as \textit{insert} and \textit{screw}. 

\begin{figure}[htbp]
  \centering
  \includegraphics[width=\linewidth]{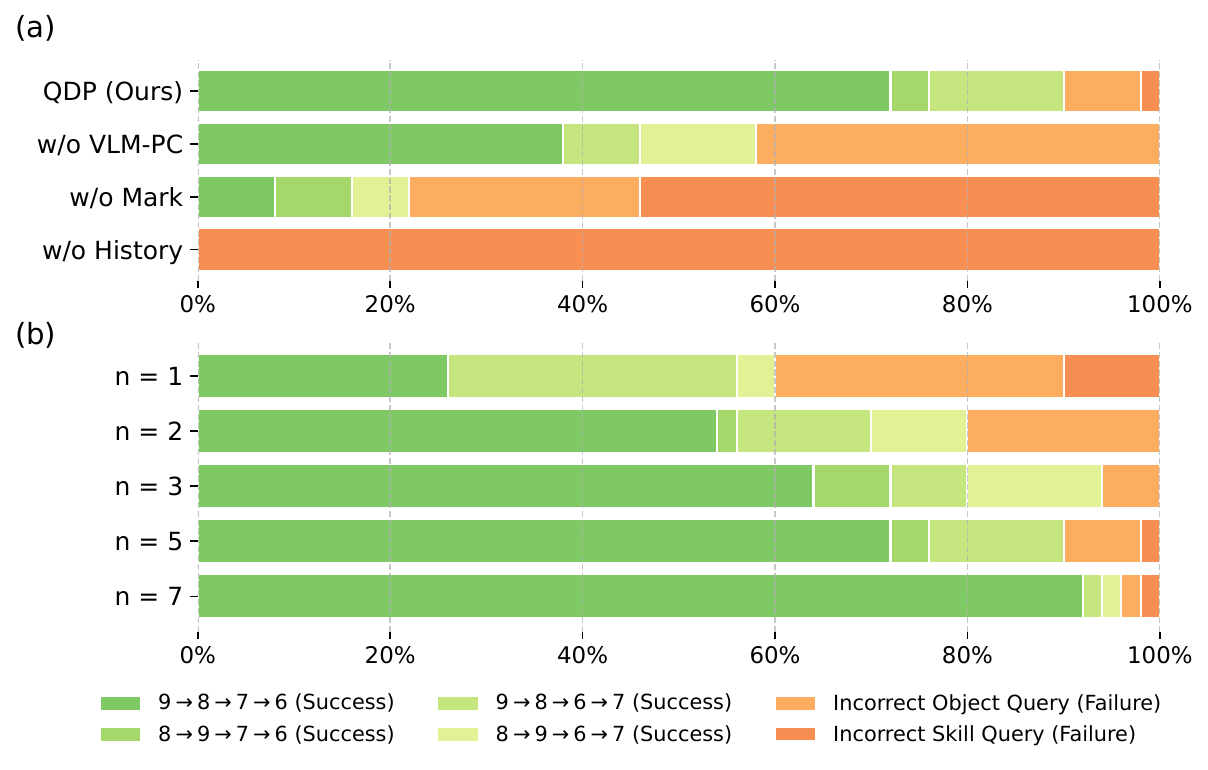}
  \caption{
  (a) Performance of different QDP high-level policy variants.
  The sequence $9 \to 8 \to 7 \to 6$ specifies the installation order of the contact points. 
  Incorrect Skill Query indicates that the installation failed because an incorrect skill type was selected. 
  Incorrect Object Query means failure occurs due to querying the wrong furniture component or Contact Point, even though the correct skill query is used.
  (b) Effect of varying history lengths on performance. 
  $n = k$ indicates that the VLM receives the current image along with the most recent $k$ steps of interaction history as input.
  }
  \label{fig:highlevel_exp}
\end{figure}

We further evaluated 3 different variants of QDP high-level policy: \textbf{QDP w/o VLM-PC}, \textbf{QDP w/o Mark}, and \textbf{QDP w/o History}.
Figure~\ref{fig:highlevel_exp}(a) shows the evaluation results for different QDP high-level policy variants; the result shows that removing specific design modules negatively impacts performance. 
Excluding the VLM-PC (\textbf{w/o VLM-PC}) significantly reduces the success rate, demonstrating the effectiveness of dynamic skill selection. Similarly, removing the SoM marks (\textbf{w/o Mark}) impairs performance; notably, more errors occur due to incorrect skill query, showing that the special reasoning capability is greatly reduced without facilitating with SoM. Finally, removing historical interactions has the most severe effect, leading to a marked increase in incorrect skill queries, emphasizing the critical role of maintaining historical context for sequential tasks.
We further analyzed the effect of varying interaction history length $n$, and the result is shown in Figure~\ref{fig:highlevel_exp}(b). 
The success rate increases rapidly as the history length $n$ increases and stabilizes after
$n \geq 3$.

\begin{figure}[htbp]
  \centering
  \includegraphics[width=1.0\linewidth]{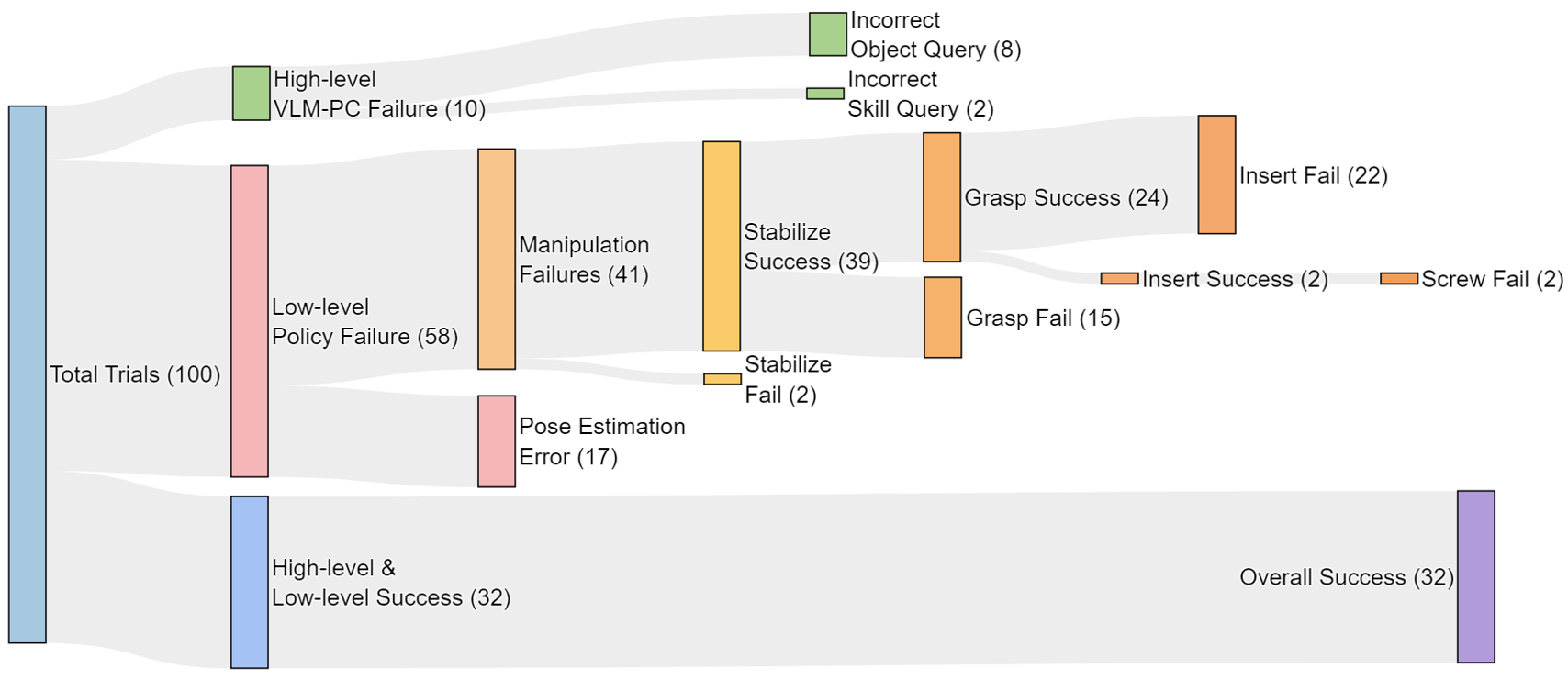}
  \caption{
  Sankey diagram illustrating the hierarchical flow of failure and success modes throughout the \textit{One Leg} process. We conduct 100 simulation experiments, with most failures occurring in pose estimation and manipulation sub-tasks such as \textit{insert} and \textit{grasp}.
  }
  \label{fig:flow}
\end{figure}

For~\qa{\textbf{RQ3}}, we evaluate the robustness of QDP under various perturbations. 
The breakdown results in the simulation are shown in Table~\ref{tab:simulation_results}(lower).
1) We found that the shape transformation (\textbf{v/ shape}) does not result in a significant drop in performance, especially in handling challenging tasks such as \textit{insert}, which aligns with our expectations. 
2) In the order variation of the final leg (\textbf{v/ order}), selecting the two holes closest to the camera often results in collisions with other assembled legs, leading to sub-task failure during the \textit{insert} phase. In contrast, switching between the two more distant holes does not result in such issues. This demonstrates that our query-centric framework generalizes across different contact points while also highlighting the importance of selecting feasible contact points in generating the assembly sequence.
3) In simulation and real-world experiments, we demonstrate the robustness of our closed-loop pose estimation and action diffusion module. Upon introducing disturbances to the states of the furniture (\textbf{v/ tabletop}), the robot effectively detects these changes, re-estimates the pose, and showcases the multi-modal capabilities of the action diffusion module. 
A qualitative example is shown in Figure~\ref{fig:robustness}(left), where the robot arm is capable of adapting to human perturbation on the tabletop when inserting the leg. 

%% file: 6_conclusions.tex
\section{CONCLUSION}
\label{sec:conclusion}

In this work, we propose QDP, a hierarchical, query-centric diffusion policy framework for long-horizon robotic assembly. By leveraging task-relevant queries, QDP effectively addresses both inter-part relational reasoning and intra-part precision control, achieving strong performance in multi-part assembly tasks. While QDP represents a significant step forward in robotic furniture assembly, it also highlights remaining challenges in achieving scalable and generalizable assembly solutions.